\documentclass[sigconf]{aamas} 

\usepackage{xcolor}
 \newcommand{\red}[1]{\textcolor{black}{#1}}

\usepackage{wrapfig}

\usepackage{float}
\usepackage[T1]{fontenc}

\usepackage{balance} 

\setcopyright{ifaamas}
\acmConference[AAMAS '23]{Proc.\@ of the 22nd International Conference
on Autonomous Agents and Multiagent Systems (AAMAS 2023)}{May 29 -- June 2, 2023}
{London, United Kingdom}{A.~Ricci, W.~Yeoh, N.~Agmon, B.~An (eds.)}
\copyrightyear{2023}
\acmYear{2023}
\acmDOI{}
\acmPrice{}
\acmISBN{}

\title[AAMAS-2023 Formatting Instructions]{Establishing Shared Query Understanding in an Open Multi-Agent System}

\author{Nikolaos Kondylidis}
\affiliation{
  \institution{Vrije Universiteit Amsterdam}
  \city{Amsterdam}
  \country{The Netherlands}}
\email{nikos.kondylidis@vu.nl}

\author{Ilaria Tiddi}
\affiliation{
  \institution{Vrije Universiteit Amsterdam}
  \city{Amsterdam}
  \country{The Netherlands}}
\email{i.tiddi@vu.nl}

\author{Annette ten Teije}
\affiliation{
  \institution{Vrije Universiteit Amsterdam}
  \city{Amsterdam}
  \country{The Netherlands}}
\email{annette.ten.teije@vu.nl}

\begin{abstract}

We propose a method that allows to develop shared understanding between two agents for the purpose of performing a task that requires cooperation.
Our method focuses on efficiently establishing successful task-oriented communication in an open multi-agent system, where the agents do not know anything about each other and can only communicate via grounded interaction.
The method aims to assist researchers that work on human-machine interaction or scenarios that require a human-in-the-loop, by defining interaction restrictions and efficiency metrics.
To that end, we point out the challenges and limitations of such a (diverse) setup, while also restrictions and requirements which aim to ensure that high task performance truthfully reflects the extent to which the agents correctly understand each other.
Furthermore, we demonstrate a use-case where our method can be applied for the task of cooperative query answering.
We design the experiments by modifying an established ontology alignment benchmark.
In this example, the agents \red{want to query each other, while} representing \red{different} databases, defined in \red{their own} ontologies \red{that contain} different and incomplete knowledge.
Grounded interaction here has the form of examples that consists of common instances, for which the agents are expected to have similar knowledge.
Our experiments demonstrate successful communication establishment under the required restrictions, and compare different agent policies that aim to solve the task in an efficient manner.

\end{abstract}


\newcommand{\BibTeX}{\rm B\kern-.05em{\sc i\kern-.025em b}\kern-.08em\TeX}

\begin{document}


\pagestyle{fancy}
\fancyhead{}


\maketitle 


\section{Introduction}


Establishing successful communication between a human and a machine is challenging.
It requires an a priori definition of all the potential scenarios under which these agents will have to communicate.
This is not always possible, especially when designing autonomous systems.
Another solution would be for the human to understand the system's operation and communicate accordingly.
This is also non trivial, as it is very challenging ``to understand, comprehend, and work with very large conceptual schemas'' \cite{villegas2013filtering}.

The adaptive ability of a system to establishing successful communication with humans in non-foreseen cases can substantially extend its (re-)usability.
This would allow for the system to learn new concepts or tasks with ad-hoc user interaction.
This is possible while not requiring the user to be familiar with the particular system, or even be a domain expert.
Communicating that via natural language is not always possible, as it would require high levels of natural language understanding and preferably in several languages.
Instead, we suggest that such an understanding can be approximated using only \red{a few examples of correct and wrong query results, provided by the user.}
Such an ability would be particularly useful in the case where \red{a person needs to query multiple databases, without having to familiarize with their respective schemas.
For example, a new Ph.D. student wants to reach out to senior researchers, i.e. people that have been members of a Program Committee of a conference. The student knows which of the people in her group have been members of such a committee and can use them to form examples to query conference organisation databases, without having to familiarise with the schema of each database.}
\red{This study aims to put together a framework with guidelines for designing communication experiments with a human-hybrid application focus.}

Different fields have studied how successful communication can be established between agents.
Studies from Ontology Alignment (OA) \cite{OM_book}, attempt to establish translations between ontologies, allowing for the systems to communicate their own concepts with each other.
These approaches require both systems to operate under formally defined ontologies and are too inefficient to be applied in real time \cite{besana2005exploiting}.
The problem has also an overlap with the Open Multi-Agent Systems (OMAS) field, where agents from a diverse population, that do not have an established communication protocol, need to learn how to communicate, only via grounded interaction. Similarly, language games \cite{steels_language_2001} also study the establishment of correct communication.
Nevertheless, in these fields, the agents do not actively attempt to communicate or explain a particular concept in an efficient manner, but instead the context of each interaction is randomly provided by the environment.
Therefore, applications of such methods are not able to represent restrictions that humans bring in the problem, i.e. not having a formal ontology for each agent, nor aim to educate or be educated by a human about a particular topic in an efficient manner.

Our research question is whether we can put together such restrictions and efficiency aspects in a framework.
This would allow the development and evaluation of intelligent systems that can learn to communicate with humans in an ad-hoc setting.
Furthermore, to minimize hardly acquired human input, this framework should allow researchers to indirectly \red{estimate} their method's efficiency, according to application-specific metrics that they can define\red{,  before running experiments with humans}.


The \red{framework} describes how two agents and an environment interact in a cycle.
The agents need to cooperate in order to perform a task on the environment.
Since at least one of the agents does not operate under a formally defined ontology, we cannot simply compare its interpretation with the intended interpretation and evaluate communication, but can only indirectly evaluate it based on the performance of a downstream task.
Neither agent can perform the task alone, so that cooperation is required and task performance reliably reflects correct understanding among the agents.
In more detail, one agent called ``Teacher'' can understand the task that needs to be performed but cannot act towards it.
Instead, it can only provide grounded examples to the other agent, the ``Student'', who needs to understand the task and perform it.
The environment is able to evaluate the Student's action, which indirectly evaluates the Student's understanding of the Teacher.
It is nevertheless important that the environment's evaluation is only presented to the Teacher, so that the Student alone cannot figure out the task over time without the Teacher's input.
This framework allows to test and compare different Teacher and Student policies, namely how they should teach or learn a new concept in terms of both task performance and efficiency   (e.g. number of interaction cycles, cognitive load, or memory demands). 

We applied this framework on the scenario where the two agents need to collectively answer a query.
Specifically, we utilize an OA dataset that provides common instances that can be used for grounded communication and concept alignments that allow the Environment to evaluate whether the two agents are talking about the same thing and correctly understand each other.
Additionally, the different ontologies contain overlapping and incomplete data, so neither agent can alone correctly answer the query successfully, and task performance is measured in terms of traditional information retrieval. 
Efficiency is principally measured in terms of number of interactions or examples that the agents exchange. 
Furthermore, we consider cognitive limitations and ask the Teacher to only provide one relevant and one irrelevant instance within each example.
Last but not least, we also attempt to represent the memory demands of the agent policies that we put forward, by defining episodic memory and working memory sizes in number of memorized episodes and variables used respectively.
\red{Please note that the cognitive load metrics used are only intended as an example of how one could define its own case-specific metrics to evaluate shared understanding using our framework and are not meant as a main reference to measure cognitive load.}
Our experiments display a use-case application of our framework that allows us to evaluate teaching and learning policies that can be applied in an OMAS.
The applied communication restrictions and efficiency metrics aim to represent that these policies can also be applied when the other agent is a human, while taking into account their estimated cognitive demand.
The experiments show that it is possible for the agents to be well understood in the described setting, using only a very limited number of interactions.
In addition, the results suggest that a robust query understanding method can be expected to perform better in a scenario where the agents have incomplete knowledge of their environment.
Finally, the experiments on the applied use-case suggest that the examples which seem to be more informative are more probable to be interpreted differently by the two agents.

In this work, \red{and specifically in Section \ref{sec:framework},} we put together i) the challenges that need to be considered when developing human-machine communication and ii) suggest guidelines on how to computationally evaluate understanding performance but also efficiency metrics into a framework that can foster research around on-the-fly human-machine communication.
Additionally, \red{Section \ref{sec:applying_framework} presents} a use-case of applying this framework on the task of cooperative query answering, where we suggest and evaluate policy agents that aim to either teach or learn \red{query-}concepts to each other in an efficient manner. \red{The queries are about roles of conference participants, i.e. authors, reviewers, program committee members, etc.}
Our experiments indicate that successful communication \red{in such a setting} is possible using only a small number of interactions.

\section{Related Work}
Several research fields are dealing with different aspects of communication establishment among agents.

\subsection{Ontology Alignment}
The field of Ontology Alignment studies how systems with different formal representations of their knowledge can communicate successfully.
An OA method traditionally takes as input a pair of ontologies and suggests concept or instance alignments \cite{OM_book,OAEI_six_years} among them.
These alignments are used as a basis for the two systems to translate knowledge between them, allowing for successful communication.
Most OA studies are very hard to be applied in our scenario. Firstly, most OA methods require complete access to both ontologies, which might not be possible to extract from humans. Secondly, the evaluation of suggested OA methods is usually performed by comparing the produced alignments with a set of gold reference alignments, which can be very expensive to get.
\red{Instead, we suggest the evaluation to be indirectly reflected on the performance of a task that requires agent collaboration.}
Last but not least, the complexity of existing methods makes their real time application almost impossible \cite{besana2005exploiting,mcneill_dynamic_2007,atencia_interaction-based_2012}.
\red{The hybrid aim of our framework requires the communication establishment methods to be evaluated in the form of trade-off between efficiency and task achievement, i.e. precision and recall), instead of only focusing on the latter.}

\paragraph{Incremental OA Through Interaction}
Other studies do not take a holistic approach that suggests all possible alignments in one go.
Instead, the ontologies are used by agents that aim to establish or to refine existing alignments through agent interaction \cite{tamma_1_laera_argumentation_2007,tamma_3_chen_using_2015,euzenat_interaction-based_2017,an_crafting_2017,aberer_start_2003,anslow_aligning_2015,besana2005exploiting}.
In some studies, the agents engage in formal negotiation procedures, the content of which is concept alignments \cite{tamma_1_laera_argumentation_2007,tamma_3_chen_using_2015}. The agents \red{have (limited) access to each other's ontologies as to subjectively} propose, reject or rebut about them, until they converge to a set of commonly agreed alignments. \red{Contrary to our framework, the agents do not strategically decide their interactions as to efficiently achieve a concrete communication outcome, while also they evaluate communication according to reference alignments instead of some downstream task performance.}
In other studies \cite{euzenat_interaction-based_2017,an_crafting_2017} \red{the agents use properties from their ontology to describe objects to one another. Eventually, they form several property alignments. In contrast, in our study, the same agent keeps on providing examples to efficiently explain one particular property to another agent. Furthermore, the examples consist of both relevant and non-relevant objects to narrow down the possible property interpretation more efficiently.}
In \cite{aberer_start_2003}, a population of agents that are grouped to represent databases in different ontologies are asked to collectively answer queries. The agents spread the queries \red{and their answers} around the rest of the population using provided concept alignments in a gossip-like way. When an agent receives again a query that it had forwarded in the past, it evaluates its translations or even replaces them, based on the translations or answers of agents that have meanwhile propagated that query. \red{Such an approach 
evaluates communication according to provided alignments and not using a task, while requiring very large communication volumes that humans could not support.}
In \cite{anslow_aligning_2015}, a population of agents that exist in a shared environment need to exchange information regarding environmental observations.
Similar to our study, their evaluation is taking place with respect to a downstream task, while they also measure communication efficiency in terms of communicated data volume. However, their work focuses on performing instance matching provided a shared schema and therefore cannot be applied when some agents are humans, as they are not expected to use the same schema.
Last but not least, research has also been performed on agents that focus on aligning only what is necessary for the current interaction, aiming to allow an efficient real time application \cite{besana2005exploiting}. Additionally, the agents store ``interaction contexts'', in order to reduce the interpretation space of future interactions based on its context. \red{In contrast to us, } the authors do not account for minimizing communication volume restrictions 
when some agents are humans. Alignments are generated following traditional structural/lexical OA metrics by comparing parts of the ontologies.

\subsection{Agent Communication Establishment}
Here, we present studies on communication establishment that can be applied on different agents, regardless of the \red{(in-)}formal system each agent operates under.

\paragraph{Communication Establishment in Open Multi-Agent Systems} O-MAS focuses on agent interaction in dynamic populations of agents.
Computationally this means that neither the agents' system, nor the task that they will need to perform, can be decided a priori.
Therefore, the agents need to learn how to communicate while making the least amount of assumptions regarding other agents.
To that end, communication or symbol alignments is established according to environment observations, \red{that is grounded communication}.

In \cite{atencia_interaction-based_2012}, the agents engage in a conversation, which both interpret according to their own automaton. Some of the communication language is shared, and the rest of the symbols are aligned after enough interactions.
The communication is considered to be successful if the agents end the conversation at the same time, which is an observable act \red{and inline with our approach, but ignores the actual outcome. Our framework requires the agents to perform an action, i.e. answer a query, to better estimate the levels of correct communication.} Experiments of that study were performed on synthetically generated automata.

In \cite{rovatsos_interaction_2003}, a set of agents learn how to interact with each other in order to achieve potentially conflicting goals.
The communicated symbols are interpreted in terms of the agent's observable actions that follow.
Agents decide what to communicate aiming to decrease interpretation entropy, but can also misuse a word as an attempt to take advantage of other agents.
\red{In this approach, agent communication is not explicitly interpreted, but indirectly affects the agent behavior, via training a Markov Decision Process that translates communication events into actions, based on environmental rewards. While this study does not focus on communication establishment, it follows our framework idea that agent learning should be based on indirect environmental rewards.}


\paragraph{Language Games}
Language games \cite{steels_language_2001} are an approach to establish successful communication in an OMAS only via grounded communication.
In this setup, a population of agents engage in pairs to play grounded referential games.
In these games, the agents develop their own vocabulary, that is interpreted in terms of object characteristics and is used to refer to a particular object in a context.
After enough games, the population of agents converges to having shared symbol interpretations, allowing them to always communicate successfully.
This work has been extended to scenarios where one agent teaches the other one how to interpret natural language questions regarding a commonly perceived scene of objects \cite{nevens-etal-2022-language}.
Additionally, there has also been a demonstration on how a human can engage with a computer to teach it a concept, by constructing examples using physical world objects \cite{nevens2019interactive}.
These studies have a different focus compared to our work. They aim to replicate the humane ability of developing a language through de-centralized peer-to-peer interactions.
The main aim of these studies is to study the evolution of natural languages, by replicating the process. 
\red{In more detail, the agents interact in multiple random scenarios of having to refer to an object in their surrounding using invented words. Eventually they converge to a commonly interpreted vocabulary that allows them to successfully communicate in any scenario.}
\red{On the other hand, our framework is designed for learning to communicate one specific concept, as efficiently as possible, by designing interaction contexts, i.e. the teaching examples, accordingly.}

\paragraph{Interactive Reinforcement Learning}
There have also been some studies that do not aim to establish human-computer communication, but use a human in the loop to create more efficient Reinforcement Learning (RL) methods.
RL studies categories of problems where the reward signal is sparse and is only provided to the agent after multiple decisions.
Such algorithms are very computationally expensive as they usually require a lot of action-reward exploration.
Interactive RL aims to incorporate human input in that process, in order to make the convergence of such approaches faster \cite{interactive_RL_survey}.
This way, interactive RL studies focus on the downstream task, and use human input as a way to improve their performance.

\section{A Framework for Developing and Evaluating Human-Hybrid Understanding}

\label{sec:framework}
\red{This section presents our framework which provides guidelines on designing and evaluating communication establishment policies in an OMAS setting, including human-computer scenarios.}
\red{Guidelines are} either in the form of restrictions, or by defining efficiency goals that should be optimized.
\red{
Our framework is designed in an abstract and modular way, allowing the experiment designer to easily define different humane aspects that best fit their scenario.}
\red{We define understanding as the product of the interpretation process. For example, two agents have a shared understanding of a query when they interpret it using equivalent classes in their own ontologies.}

\begin{figure}
\centering
\includegraphics[width=0.35\textwidth]{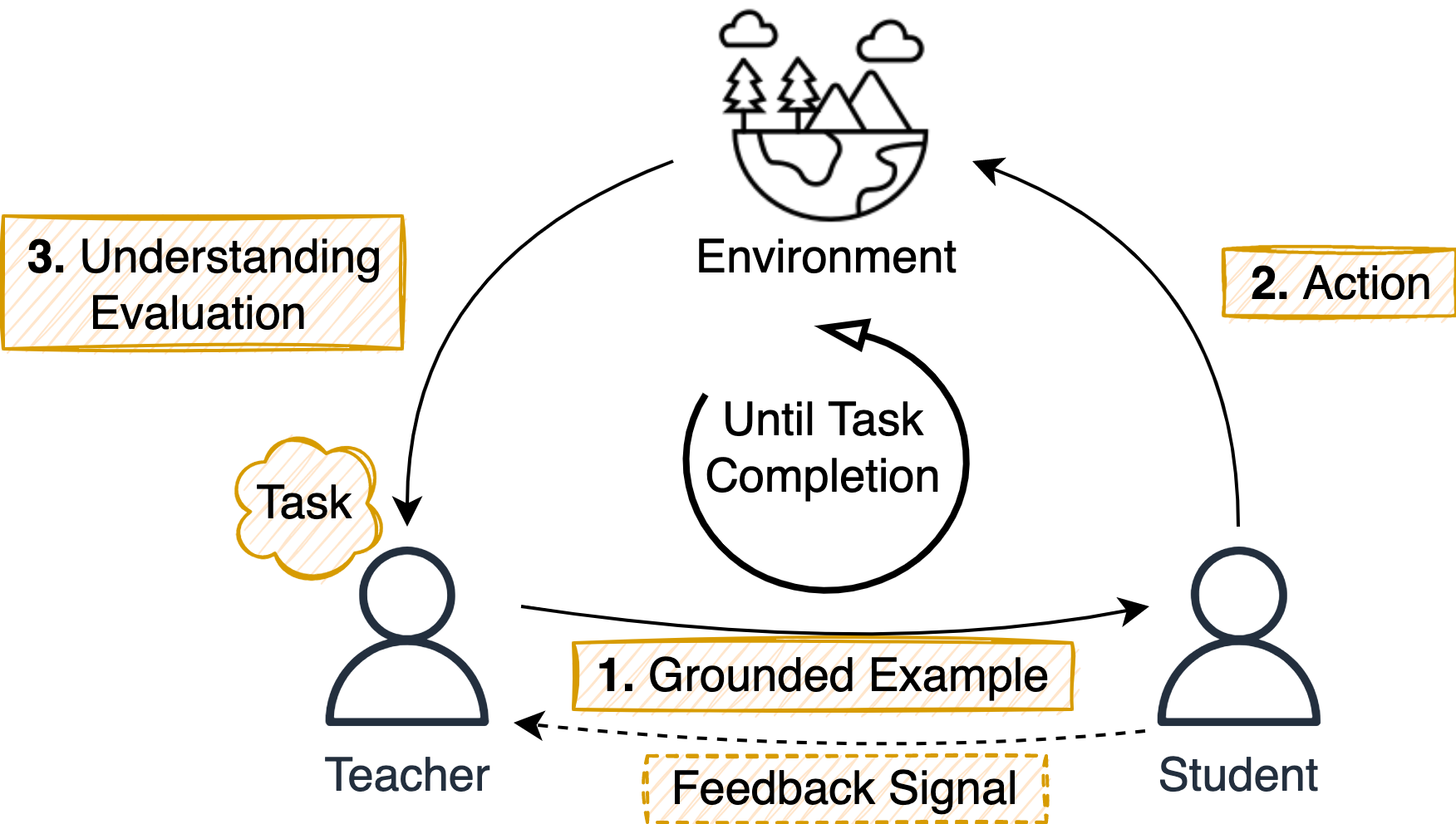}
\caption{\label{fig:framework_figure}
The task-oriented information flow cycle for shared understanding. One agent, i.e. the Teacher, understands the task that has to be performed, while the other agent i.e. the Student, does not, while the latter is the only one who can perform the task.
The agents engage in an iterative loop: 1. Grounded Example: the Teacher attempts to explain the task at hand to the Student providing an example; 2. Action: the Student follows by attempting to perform the task; 3. Understanding Evaluation: the Teacher \red{is informed by the Environment regarding the Student's performance on the task, which reflects its understanding of the task.} }
\vspace{-.3cm}
\end{figure}

\paragraph{The Process of Defining the Problem Computationally}
Our framework proposes a way of establishing successful understanding between two agents that is indirectly evaluated with respect to a downstream task.
The application of our framework follows a process of three steps so that it can be computationally studied.

\textbf{First}, one needs to define what consists a grounded interaction in their scenario.
These are objects that both agents \red{know and can refer to} during an interaction, while each of them has their own way of perceiving them.
Therefore, a set of common objects must be identified among the agents. In our example, these are the \red{researchers of the student's group}.

\textbf{Second}, it is important that the downstream task cannot be solved by either agent independently and that there is no conflict of interests.
To that end, the agents cannot have complete information about the environment, requiring input from both of them to successfully solve the downstream task.
As a consequence, the agents can either be designed to not be aware of all the objects of the environment or at least not to know all their properties.
In our example, while the \red{student can find all researchers, she does not know who are senior researchers}.
At the same time, \red{the database agent does not know what the student is looking for.}

\textbf{Third}, the cognitive load of the involved agents should be somehow computationally approximated so that it can then be minimized.
\red{This should be defined by the researcher, having in mind the target group of the end user that will use their application.}
\red{It can} be in the form of number of interactions, or additionally Episodic or Working memory demands, depending on how we assume the human to behave when interacting with the computer.

Once we have these, we suggest that the agents interact in a task-oriented loop that allows for incremental shared understanding development and its evaluation.
Figure \ref{fig:framework_figure} illustrates the abstract \red{share understanding establishment cycle as described by our framework, while Figure \ref{fig:understanding_game_flow} instantiates it for our specific example use-case, as described in Section \ref{sec:applying_framework}}.
Initially, one agent, i.e. \red{the Ph.D. student}, is provided with a task from the environment, i.e. \red{find senior researchers}. \red{The task can also come from the Teacher, but only the environment can fully evaluate its completion.}
Then, the agents engage in a 3-step cycle.
\red{In the first step,} the Teacher, i.e. \red{the Ph.D. student}, provides some objects as examples using \red{people from her group}.
\red{In the second step}, the Student, i.e. \red{agent database}, \red{provides what it currently interprets as} relevant objects from its complete set of objects, i.e. \red{researchers}.
\red{In the third step}, the Teacher observes the Student's object selection, i.e. \red{list of assumed senior researchers}, and the Environment, i.e. \red{the supervisors}, acts as an object selection evaluation.
By repeating this cycle, the evaluation\red{, i.e. query performance,} should improve over time.

\subsection{\red{The Process of Defining the Problem Computationally}}
\label{subsec:defining_the_problem_computationally}
Establishing successful communication between agents can be challenging because they are designed to have different goals or beliefs. These aspects shape how the agent perceives the world and give them a potentially unique perspective. A simple way to computationally define different perspectives is by assuming that the agents are operating using different ontologies. Provided that these agents exist in a similar or the same environment, we expect their ontologies to have some overlap both in terms of concepts and knowledge, while also their ontologies are expected to be incomplete. \red{Following, we will present how we apply the three steps of defining our problem computationally for a scenario that agents operate under different ontologies.}

\paragraph{\red{Step 1: Defining Communication Restrictions.}} 
Following traditional OMAS restrictions, the agents share limited to none common language and can only communicate via \textbf{grounded interaction}, which for example can have the form of referring to common world objects of their shared surrounding.
\red{Besides being a good fit for an OMAS setting, it is inspired by the Language Games studies \cite{steels_language_2001}, which are an attempt to mimic how a population of humans learn to interpret a vocabulary similarly.}
Nevertheless, they have in mind the task that needs to be performed and are aware of possible communication action interpretations.
Additionally, \red{given the participation of human agents, the communicated examples should be} \textbf{concise} so that their comprehension is not very laborsome.

\paragraph{\red{Step 2:} Defining Cooperation Restrictions.}
It is very challenging to evaluate communication in an OMAS setup when some of the agents are humans.
One cannot simply compare the interpretation of the receiver and the intended interpretation of the sender, since there is no agent that understands both system representations.
Therefore, we suggest that the agents need to perform a task on the observable environment, that \textbf{requires their cooperation}.
This way the environment can act as a communication evaluator, since task performance depends on cooperation which requires successful communication.
In order to make sure that this is the case, it is important that neither agent can perform the task by independently interacting with the environment.
Finally, we assume that there is \textbf{no conflict of interest} among the agents, so that they have no reason to misuse communication signals for their own interest.

\paragraph{\red{Step 3: Defining} Efficiency Aspects}
\label{subsec:efficiency_aspects}
Since some involved agents are humans, it is important to have efficient interactions.
To that end, efficiency must be somehow defined, so that its evaluation can at least be approximated.
While these definitions depend on the specific application, we suggest two type of efficiency evaluations to be taken into account.
First, it is important to minimize \textbf{interaction time}, which can be approximated by the number of interactions needed.
Second, the \textbf{agents' effort during the complete sequence of interactions} should also be taken into account, until they quit or perform the task successfully. This can for example be represented roughly in terms of episodic and working memory demands, as this can be approximated by the number of \red{examples} that the agents had to memorize or by the number of variables that they need to keep track of, respectively.

\section{An Applied Use Case on Artificial Agents}
\label{sec:applying_framework}
In this section, we \red{present an example application of our Framework in a scenario where two agents query each other about conference participants and their roles.} 
Although we apply it on artificial agents, they behave as if they interact with humans, respecting all communication and efficiency restrictions.

\paragraph{An Example Use-Case}
\red{
We consider a human and an artificial agent (representing a database) querying each other. Whether the human wants to query the database, or the database agent aims to extend its knowledge, the aim is to allow knowledge exchange without the human having to learn the schema of the database. 
Let us assume a new Ph.D. student who want to reach out to a broader circle of senior researchers.
She knows some researchers from her group, and is able to distinguish who is senior and who is not.
While there are several conference organisation databases she can query, it is hard for her to define what a senior researcher is, let alone to form such a concept under the different schemas of all the databases.
Instead, she can use the people of her group to form examples of senior and junior researchers.
She provides such examples to the artificial agent, who tries to understand the query and returns results based on its knowledge. Since the student does not know who are the senior researchers, she is unable to evaluate the results herself.
She can still ask her supervisors to evaluate the results, as this takes less of their time compared to coming up with the list themselves.
In this case, the student acts as a Teacher, the database agent as a Student and the supervisors as the Environment.
}

\subsection{Applying Our Framework}
In this section, we will go through the three steps of defining our framework computationally, as described in Section \ref{subsec:defining_the_problem_computationally}.

\paragraph{\red{Step 1: Defining} Communication Restrictions}
\textbf{Communicating over common objects} is achieved by referring to \red{researchers} that both agents, i.e. \red{the Ph.D. student} and the \red{database agent}, are aware of. 
\red{Furthermore,} \textbf{concise interaction} is achieved by \red{asking the Ph.D. student} to only provide examples that consist of \red{one junior and one senior researcher} at a time. 

\paragraph{\red{Step 2: Defining} Cooperation Restrictions}
\textbf{Successful communication and cooperation is required} in our example. The \red{student} cannot find \red{senior researchers} alone, as \red{she would have to waste a lot of her supervisors' time}. At the same time, the \red{database agent} alone cannot \red{automatically be updated about new roles of researchers in conferences}. It is also clear why there is no conflict of interest, as neither agent has any incentive to mislead the other.

\red{
\paragraph{\red{Step 3: Defining Efficiency Aspects}}
Interaction time is measured by the number of examples exchanged between the agents and cognitive load is measured based on the estimated memory volume needs of the participating agents (cf. Section \ref{subsec:evaluation_metrics} for details).
}

\begin{figure}
\centering
\includegraphics[width=0.47\textwidth]{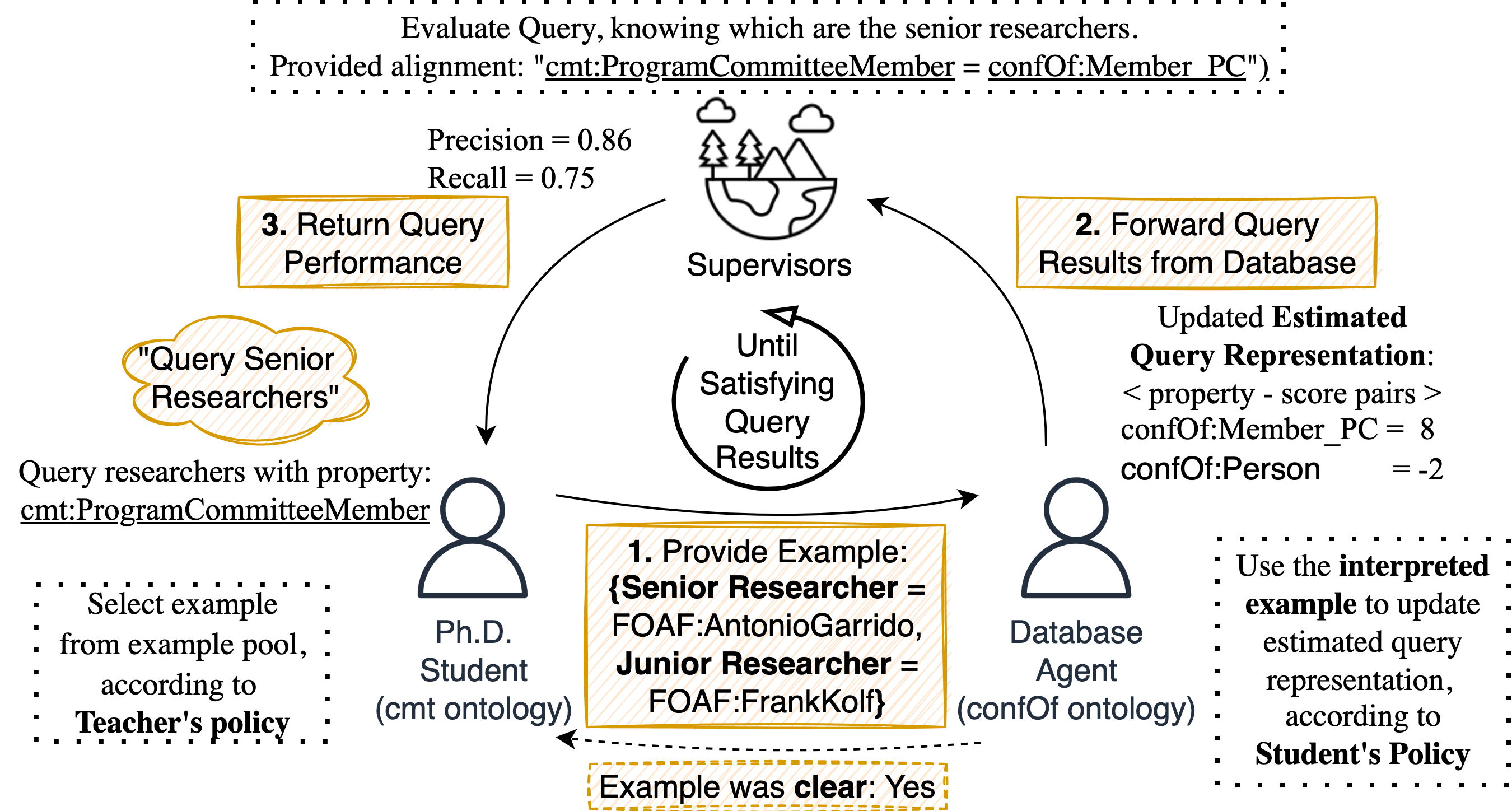}
\caption{\label{fig:understanding_game_flow}
\red{The shared understanding cycle instantiated for our example. Step (1) \textit{Provide example}: 
the Ph.D. student (acting as Teacher) wants to query the agent database (acting as Student) about senior researchers (``cmt:ProgramCommitteeMember''). Examples (URIs for senior and junior researcher) both the Ph.D. student and the database agent are aware of those common objects. Step (2) \textit{Forward Query}: 
the agent database needs to interpret the example and apply its \textbf{Student policy} to update its current \textbf{estimated query representation}, and forward the query results from its database to the supervisors (Environment). Step (3) \textit{Return Query Performance}: evaluation in terms of Precision and Recall, and inform their student how well the agent database understands her query.}}
\vspace{-.3cm}
\end{figure}

\begin{figure}
\centering
\includegraphics[width=0.47\textwidth]{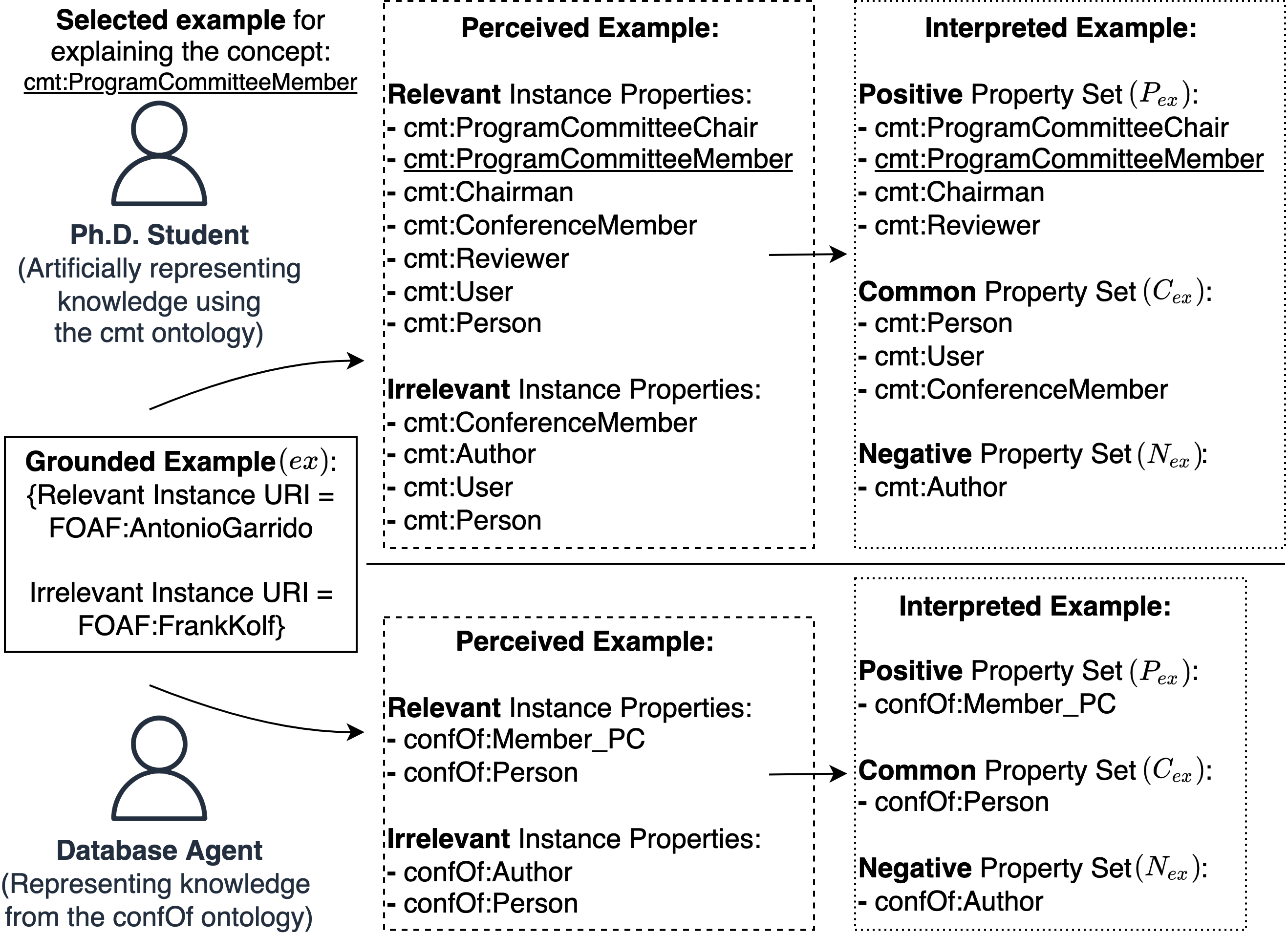}
\caption{\label{fig:example_interpretation} \red{Example of how two agents can perceive the same grounded example (URIs) differently according to their ontologies. The URIs are grounded: both ontologies refer to the same ``entities''.}}
\vspace{-.3cm}
\end{figure}

\subsection{Task Definition}
The agent interaction and the task they need to solve are presented in Figure \ref{fig:understanding_game_flow}.
The Teacher needs to explain the query to the Student who can then answer it.
\red{The Teacher's knowledge is artificially represented by another database, i.e. 
the cmt ontology in our example.
The Student's database includes knowledge of the confOf ontology.
Communication has the form of examples of world objects, i.e. researchers, that both ontologies are aware of.}
The queries are always about world objects that share some properties.
Furthermore, the agents are not aware of all the objects, nor of all the properties of the objects that they know.
Thus, only the Environment, \red{i.e. the supervisors}, is up to date and knows the complete answer set of the query. 
Our task focuses on evaluating the answers of the Student.
The evaluation of the Student's answers is not provided to the Student, as this could allow the Student to understand the query alone and make the agent communication dispensable.

\paragraph{Subjective Example Interpretation}
While the examples consist of objects that both agents are aware of and the identity of the relevant object is known, the examples are perceived differently by each agent, \red{as presented in Figure} \ref{fig:example_interpretation}.
The agents \red{independently} use their own databases to perceive the objects\red{, i.e. retrieve their properties}.
Then, they interpret the examples by forming three property sets.
The \textbf{Positive} and \textbf{Negative} property sets ($P_{ex}$ and $N_{ex}$) of an example ($ex$) are the ones that consist of properties only owned by the relevant and irrelevant object respectively.
Additionally, the agents can also calculate the set of \textbf{Common} properties ($C_{ex}$) of the two objects.
\red{The Teacher, i.e. Ph.D. student, communicates examples that contain the property it wants to explain in their Positive property set.}
\red{
In case they example interpretation of the Student, i.e. database agent, has an empty Positive property set, then this example is \textbf{unclear} for the Student.
A clear example, according to the Student, does not necessarily mean that it was understood as the Teacher intended.
This would be the case if the same communicated example presented in Figure \ref{fig:example_interpretation} was used to explain the concept ``cmt:Chairman'', or ``cmt:Reviewer''.}
Therefore, due to different databases, the Teacher might need to provide several examples until the Student can fairly approximate the query at hand.

\subsection{Teacher Policies}
Provided the property of the query, the Teacher creates an \textbf{example pool}, with all examples i.e. object pairs, where the query property \red{is contained} in their Positive property set.
The Teacher has to decide which example to provide next.
This is performed by the Teacher's policy which scores the examples in that pool, based on the history of examples and the Student's feedback. \red{The Student's feedback is used to communicate when an example was unclear, i.e. has an empty Positive property set according to the Student. The Teacher cannot foresee these cases, but can instead memorize these examples and not present them again by removing them from the example pool.}

\paragraph{Random Teacher Policy \red{(\cite{an_crafting_2017})}}
One option is for the agent to follow a random teaching policy, where examples are randomly sampled from the pool, with replacement. \red{This agent policy is an adaptation of the proposed method in \cite{an_crafting_2017}, for our task, where the objects that the agents describe is chosen randomly.}

\paragraph{Property-Based Teacher Policy}
This policy increases the score of an example according to the properties that it aims to exclude from query interpretation candidates.
We define the \textbf{weight} ($w_p$) of each property $p$ as the number of objects that have this property, compared to the total number of objects in the database (its normalized frequency).
Then, we define the score of each example ($score_ex$) as:
\[ score_{ex} = \sum^{\forall p \in (N_{ex} \cup C_{ex} )} w_p - \sum^{\forall p \in P_{ex}} w_p \]

\subsection{Student Policies}
The Student's policy provides instructions on how to translate the incoming stream of examples to an \textbf{estimated query representation}, which is then used to provide results to the Environment.
It is common knowledge that the query is about a set of objects that share some property.
Thus, the estimated query representation consists of a set of property-score pairs. \red{The Student policies only make use of the Positive and Common property sets of an example.}
We will now present the two policies that we are comparing. 

\paragraph{Logic-Based Student Policy}
This policy requires the Student to memorize the provided examples, and then apply Formal Concept Analysis (FCA) \cite{ganter2012formal} to deduce which properties are implied by the query.
FCA can be used to derive property relations, provided a set of objects and their relations.
Therefore, we treat each example as an individual \textit{object}, and the Positive property set of that episode ($P_{ex}$), as the properties of this \textit{object}.
The Student memorizes a set of unique Positive property sets of examples.
The query is also represented as a pseudo property of these example \textit{objects}.
This way, FCA provides us with relations between the query and the properties of the agent's ontology.
Only the properties that can be induced from the query pseudo property are found on the estimated query representation set, and they all have the same score, so that they affect the produced results in the same way.

\paragraph{Frequency-Based Student Policy}
The Frequency-based policy does not require the student to memorize any example, but only to maintain and update scores per observed property.
Specifically, the agent starts with an empty set of property-score pairs, as its estimated query representation.
For every new example, the Student makes use of its Positive and Common property sets.
In case any of these properties is not in the estimated query representation, they are added with an initial score of 0.
Then, the score of each property on the Positive or Common property sets of the example is increased or decreased by 1, respectively.

\paragraph{Providing Query Results}
When asked, the Student uses its estimated query representation to rank all objects in its database and provides them to the Environment. The score of each object is equal to the summation of the scores of its properties, as they are calculated in the estimated query representation. The agent only returns objects with positive scores.

\section{Experiments}
In this section, we  present the experiments that we performed.
Our code is publicly available\footnote{https://github.com/kondilidisn/shared\_query\_understanding}, ensuring reproducibility of our work.

\subsection{Dataset}
For our experiments, we used the dataset of the Ontology Alignment for Query Answering (OA4QA) track \cite{solimando2014evaluating}.
The dataset contains 7 populated ontologies, that have on average 817 instances belonging to 17 classes.
The ontologies describe an overlapping set of conference venues, with information such as people's role, types of submitted papers, etc. About 259 class alignments are present over the 21 ontology pairs.
This dataset is a populated version of the popular dataset used on the Conference track of the well established Ontology Alignment Evaluation Initiative\footnote{http://oaei.ontologymatching.org/2022/conference/index.html}, although we only use the instances and their classes.

\paragraph{Utilizing the Dataset}
We only focus on the instances and their classes from each ontology, that we use as world objects and their properties respectively.
This dataset is a very good fit for our scenario, as it contains several class alignments.
These can be used as equivalent object properties across agent databases, over which we can perform query understanding experiments.
Furthermore, since the ontologies describe a set of overlapping venues, they contain different and incomplete knowledge about their environment.
This way, we ensure that cooperation is required, since no agent can alone provide all answers in any query understanding experiments, as the complete answer set is always distributed among both agents.

\paragraph{Crafting Query Understanding Experiments}
Every provided class alignment between two ontologies can be turned into a query understanding experiment.
Specifically, each alignment can generate two experiments, since the agent that represents one of the two ontologies can either participate as a Teacher or as a Student.
In order for the alignment to be put to use, both ontologies need to have at least one instance of their corresponding aligned class in the set of common instances between the two ontologies.

\paragraph{The Simple Dataset}
As previously described, the number of common instances across ontology pairs has a direct effect of how many of the provided class alignments can be used as query understanding experiments.
Instance alignment across ontologies is not a problem that we aim to address in this paper.
Instead, we aim to identify common instances across ontologies, and create two different datasets accordingly. In the first dataset, denoted as Simple, two instances are the same only if they have exactly the same Unique Resource Identifier (URI). This dataset has on average 303 common instances across pairs of ontologies, and can be used for running 126 experiments.

\paragraph{The Extended Dataset}
The second dataset, denoted as Extended, extends the first dataset by making three more assumptions on whether two instances are the same across ontologies.
The first extra assumption suggests that two instances with the same name but different namespace are the same.
The second extra assumption suggests that two instances are the same, if the are related in the same way, i.e. equivalent relation, to another instance that is known to be the same.
This would be formalized as o1:x == o2:y if (common\_entity, common\_relation, o1:x) AND (common\_entity, common\_relation, o2:y). Last but not least, we also mined some instance alignments by executing complex queries. For example, in one ontology, the relation \textit{o1:writes} can connect a \textit{o1:Person} with either a \textit{o1:Paper}, or with a \textit{o1:Review}. In the other ontology, \textit{o2:has\_written} only relates \textit{o2:Person} with \textit{o2:Paper}. 
In this case, we first retrieved all \textit{o1:Person} that were related with instance \textit{o1:X}, with the relation \textit{o1:writes}.
Then, we filtered out all \textit{o1:X} that were not \textit{o1:Paper}, and finally applied our second assumption as described before.
This dataset has 482 common instances across ontology pairs on average, and allowed the execution of 226 different experiments.

\paragraph{Aggregating Evaluations Over Multiple Queries and Experiments}
Each experiment represents the behavior of the combination of Teacher's and Student's policy for a specific dataset.
This is calculated by averaging the recorded evaluation metric values over all the query understanding experiments that can be performed on that dataset.
Additionally, some agent policies are non-deterministic, which can be the case because scoring ties are resolved randomly.
For this reason, we repeat every experiment of a specific query 10 times, and report average values of the evaluation metrics.

\subsection{Evaluation Metrics}
\label{subsec:evaluation_metrics}
Our evaluation focuses both on how well the agents understand each other, i.e. reflected by the task performance, while also the efficiency costs of each approach.

\paragraph{Task Performance}
The task that the agents are trying to collectively solve is query answering. Therefore, their performance is measured in terms of traditional information retrieval terms, i.e. Precision and Recall.

\paragraph{Efficiency Costs}
The interaction time is measured as the number of examples that were communicated between the agents, i.e. \#examples.
Furthermore, we use the notions of Episodic Memory, and Working Memory, to approximate the agents' cognitive effort throughout the complete sequence of interactions.
The Episodic memory is defined as the number of episodes that the agent has memorized.
These are the \red{unclear} examples for the Teacher and the clear examples for the Logic-based Student.
The Working memory is equal to the number of variables that the agent policy is using in order to operate, i.e. the property scores. \red{The terms Episodic and Working memory here are used as an example of how one can measure cognitive effort.}

\subsection{Experiment Results and Discussion}
\begin{figure}
\centering
\includegraphics[width=0.49\textwidth]{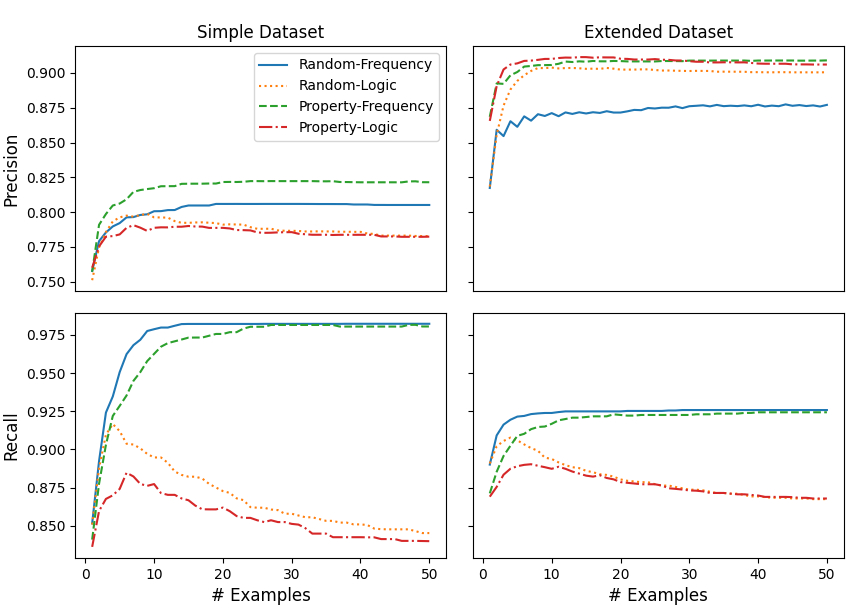}
\caption{\label{fig:performance} Performance evaluation of the proposed agent policies on the two datasets. The legends denote the Teacher and Student applied policy respectively, separated by `-'.}
\vspace{-0.3cm}
\end{figure}

Figure \ref{fig:performance} depicts how the suggested agent policies behave in terms of task performance, with respect to the number of examples exchanged.
Recall that the evaluation takes place on the query results of the Student to the environment, as a way of indirectly measuring how well the Student understands the Teacher\red{'s query}. 
An important takeaway is that in any combination of selected policies, the agents achieve more than 80\% Precision and around 90\% Recall, within only a handful of provided examples.
Nevertheless, it is interesting to note that when the Student is following the Logic-based policy, Recall starts to decline after only a few examples.
This suggests that the Logic-based policy is not robust enough for scenarios with ontologies that contain incomplete and partial knowledge.
Similar behavior is also observed on the Precision of the Logic-based Student policies, but only on the small dataset.
Furthermore, when comparing the applications of Frequency-based Student policies, we can see that a Teacher that applies the Property-based policy helps achieving a good agent understanding in fewer examples. 
Moreover, it is also interesting to note that only in the large dataset, a Student following a Frequency-based policy 
 and taught by a Teacher with Random policy, is under-performing in terms of Precision compared to all other policy combinations.
Besides, all policy combinations achieve higher Precision on the extended dataset.
Additionally, the Frequency-based Student policies show a drop of Recall on the Extended dataset.

\paragraph{Efficiency Metrics Performance}
When the Student applies the Frequency-based policy, its working memory is around 5,5 and 4,6 on the small and the extended dataset respectively, and equal to 0 for the Logic-based policy. This number reflects the average number of properties on the estimated query representation. It seems that on average, there are less potential candidate properties over which the examples are interpreted on the Extended dataset, which also explains the higher average performance of all experiments there.
Complementary, the Student only uses its Episodic memory when following the Logic-based policy, in which case the agent memorizes around 1.6\% examples on average in all experiments. This suggests that there are equally few distinct useful examples to memorize their properties, across all experiments. Last but not least, the Episodic memory of the Teacher, which reflects the number of \red{unclear} examples found, is much smaller on the extended dataset, ranging from 2\% to 4\%, compared to the Simple dataset where it ranges from 9\% to 12\%. This suggests that maybe the knowledge overlap across the ontologies is bigger on the extended dataset on average. Moreover, this shows that the examples which are expected to be more informative are more probable to be perceived differently by the two agents in our use-case. Additionally, when the Teacher is following the Property-based policy, it ends up memorizing on average around 2\% to 3\% more \red{unclear} examples. One can therefore say that the Teacher is more likely to select an example that contains instances with different information across ontologies, when attempting to find more informative examples.

\paragraph{Discussion}
The experiments exhibit that our framework can be applied for incrementally establishing a shared understanding between the agents.
In more detail, it is shown that a few concise examples are enough for the agents to develop a sufficiently good enough understanding of each other.
Moreover, it seems that the cognitive load, i.e. Episodic and Working memory demands, were not too high, according to our \red{example} estimations.

\section{Conclusion and Future Work}

We have described a framework that \red{can assist researchers in} developing agent policies that  establish successful communication in an OMAS setting, like in a human-hybrid scenario.
Additionally, we provide an application example the framework on a collaborative query answering task.
Our experiments display satisfactory levels of communication for the task at hand, while doing so in a very small number of interactions.

There are several ways to extend the presented study.
\red{
First, we are already working on demonstrating the generalizability of the framework by applying it in different studies of communication establishment including most studies presented in our related work. 
Second, we will apply the framework on more challenging tasks, which will require improving the presented agent policies with particular focus on using both the Student's feedback signal and its current level of understanding.
Such tasks can either be done by focusing on collaborative query answering using larger ontologies with complex alignments, or use tasks that include actual interaction with an environment besides information exchange.
Third, we aim to focus on the human participant aspect of our framework.
We can for example use relevant literature to apply more sophisticated methods to estimate cognitive load, or involve human participants and ask them to self-report the overall effort required for one interaction or the complete experiment.
Besides improving the estimation of human participants' effort, we plan designing artificial agents that 
take into account the behavior of a target group of end-users, requiring for example the design of robust agent policies that can interact with irrational human input.
}

\subsection*{Acknowledgments}

\begin{wrapfigure}{l}{0.025\textwidth}
\includegraphics[trim={0cm 10cm 6cm 25cm},width=0.05\textwidth]{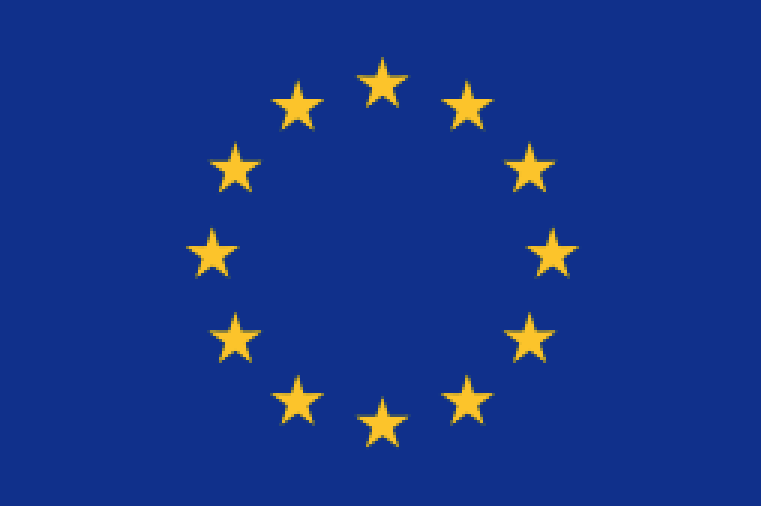}
\end{wrapfigure}
This work was supported by ``MUHAI - Meaning and Understanding in Human-centric Artificial Intelligence'' project, funded by the European Union's Horizon 2020 research and innovation program under grant agreement No 951846.
We also want to heartily thank prof. Frank van Harmelen for his support, guidance and for being a continuous source of inspiration. 

\newpage


\newpage
\bibliographystyle{ACM-Reference-Format} 
\bibliography{sample}


\end{document}